\documentclass[11pt, a4paper, logo, copyright, nonumbering]{xiaomi}

\usepackage[authoryear, sort&compress, round]{natbib}
\usepackage{dblfloatfix}
\usepackage[normalem]{ulem}
\usepackage{caption}
\usepackage{xspace}
\usepackage{pifont}
\usepackage{multirow}
\usepackage{tcolorbox}
\usepackage{xltabular}
\usepackage{longtable}
\usepackage{hyperref}
\usepackage{wrapfig}
\usepackage{graphicx}

\usepackage{amsfonts}
\usepackage{amsmath}
\usepackage{amssymb}
\usepackage{lineno}
\usepackage{multirow}
\usepackage{adjustbox}

\usepackage[bottom]{footmisc}

\usepackage{CJKutf8}
\usepackage{subfigure}
\usepackage{setspace}
\usepackage{makecell}
\usepackage{graphicx}
\usepackage{subcaption}
\usepackage{multicol}

\usepackage{afterpage}

\hypersetup{
  pdftitle={MOPD: Multi-Teacher On-Policy Distillation for Capability Integration in LLM Post-Training},
  pdfauthor={Wenhan Ma and Jianyu Wei and Liang Zhao and Hailin Zhang and Bangjun Xiao and Lei Li and Qibin Yang and Bofei Gao and Yudong Wang and Rang Li and Jinhao Dong and Zhifang Sui and Fuli Luo},
  pdfsubject={Multi-teacher on-policy distillation for capability integration in LLM post-training},
  pdfkeywords={on-policy distillation, post-training, reinforcement learning, capability integration, large language models},
}

\definecolor{xiaomiorange}{HTML}{FF6901}

\newcommand{\MOPD}{\text{MOPD}}
\newcommand{\MixRL}{\text{Mix-RL}}
\newcommand{\RFT}{\text{Off-Policy Finetune}}
\newcommand{\PM}{\text{Param-Merge}}
\newcommand{\SeqRL}{\text{Cascade RL}}
\newcommand{\sg}{\mathrm{sg}}

\title{\centering MOPD: Multi-Teacher On-Policy Distillation for Capability Integration in LLM Post-Training}

\makeatletter
\renewcommand{\author}[1]{\gdef\@author{\Authfont #1}}
\makeatother
\renewcommand{\thefootnote}{\fnsymbol{footnote}}
\author{%
  \bfseries Wenhan Ma$^{1,2}$\footnote{Work done during an internship at Xiaomi.} \quad
  Jianyu Wei$^{2}$ \quad Liang Zhao$^{2}$ \quad Hailin Zhang$^{2}$ \quad Bangjun Xiao$^{1,2}$ \\
  \bfseries Lei Li$^{2,3}$ \quad Qibin Yang$^{1,2}$ \quad Bofei Gao$^{1,2}$ \quad Yudong Wang$^{1,2}$ \\
  \bfseries Rang Li$^{1,2}$ \quad Jinhao Dong$^{4,2}$ \quad
  Zhifang Sui$^{1}$\footnote{Co-corresponding authors.} \quad
  Fuli Luo$^{2}$\footnotemark[2] \\[5pt]
  \normalfont
  $^{1}$Peking University \quad $^{2}$LLM Core, Xiaomi \\
  $^{3}$University of Hong Kong \quad $^{4}$Renmin University of China
}

\begin{abstract}
Modern large language models (LLMs) rely on reinforcement learning during post-training to push specific capabilities, yet integrating multiple capabilities into one model remains hard. Existing methods, such as \RFT{} and \MixRL{}, are either inefficient or lose performance. In this work, we propose \textbf{Multi-teacher On-Policy Distillation} (\MOPD{}), a post-training paradigm for combining the capabilities of multiple domain RL teachers: we first run per-domain specialised RL to obtain a set of domain teachers, then distill these teachers into the student on its own rollouts. This eliminates exposure bias and provides a dense optimization signal. On Qwen3-30B-A3B, \MOPD{} outperforms \MixRL{}, \SeqRL{}, \RFT{}, and \PM{} baselines, inheriting nearly all of each teacher's capability. \MOPD{} also enables parallel, independent development of domain teachers, removing the cross-domain coupling typical of multi-domain post-training. \MOPD{} has been deployed in the post-training of MiMo-V2-Flash, an industrial-scale frontier model, demonstrating its practical value for capability integration in frontier-scale LLMs.
\end{abstract}

\begin{document}
\maketitle
\renewcommand{\thefootnote}{\arabic{footnote}}

\section{Introduction}
\label{sec:intro}

Reinforcement learning (RL) has emerged as an important method for improving the capabilities of large language models (LLMs)~\citep{ouyang2022instructgpt, schulman2017ppo, shao2024deepseekmath}. Different task domains rely on different RL pipelines. Mathematical reasoning is now trained with verifiable-answer RL~\citep{shao2024deepseekmath, guo2025deepseekr1}; software engineering with agent-style RL in executable sandboxes~\citep{wei2025swerl, jain2025r2e}; instruction following and creative writing with rubric-based RL~\citep{ouyang2022instructgpt}; and search-oriented agents with web-environment RL~\citep{nakano2021webgpt, jin2025searchr1}. Each pipeline reliably improves the model's capability on its target domain. However, what we ultimately want is a single model that performs well across all of these domains. Yet building such a model remains an open problem in modern LLM post-training.

\begin{table}[!t]
  \centering
  \caption{Capability-integration paradigms compared across three axes that matter in practice. \MOPD{} is the only method that simultaneously attains dense optimization, on-policy training, and a parallelisable pipeline.}
  \label{tab:headline}
  \small
  \renewcommand{\arraystretch}{1.15}
  \setlength{\tabcolsep}{3.5pt}
  \begin{tabular}{l c c c}
    \toprule
    & Dense optimization & On-policy & Parallelisable \\
    \midrule
    \PM{}                   & \textcolor{red!80!black}{$\times$}           & \textcolor{gray}{\textemdash}            & \textcolor{green!60!black}{$\checkmark$} \\
    \RFT{}                  & \textcolor{green!60!black}{$\checkmark$}    & \textcolor{red!80!black}{$\times$}       & \textcolor{green!60!black}{$\checkmark$} \\
    \MixRL{}                & \textcolor{red!80!black}{$\times$}           & \textcolor{green!60!black}{$\checkmark$} & \textcolor{red!80!black}{$\times$} \\
    \SeqRL{}                & \textcolor{red!80!black}{$\times$}           & \textcolor{green!60!black}{$\checkmark$} & \textcolor{red!80!black}{$\times$} \\
    \textbf{\MOPD{} (ours)} & \textcolor{green!60!black}{$\checkmark$}     & \textcolor{green!60!black}{$\checkmark$} & \textcolor{green!60!black}{$\checkmark$} \\
    \bottomrule
  \end{tabular}
\end{table}

Prior approaches to this capability-integration problem fall into four families (Table~\ref{tab:headline}). (1) \MixRL{}~\citep{qwen2025qwen3} pools prompts from every domain into one dataset and runs joint RL, but cross-domain training signals interfere, producing the see-saw effect~\citep{standley2020seesaw} and leaving the joint model below per-domain teachers. (2) \SeqRL{}~\citep{wang2026nemotroncascadescalingcascadedreinforcement} trains domains sequentially, so earlier-stage capabilities may decay as later stages progress and the long overall run amplifies stability risks. (3) \RFT{}~\citep{deepseek2025v32} trains per-domain RL teachers and then fine-tunes the student on their rollouts via SFT; this off-policy supervision induces the canonical exposure bias~\citep{ranzato2016exposure}. (4) \PM{}~\citep{wortsman2022soups, ilharco2023arithmetic} averages teacher weights or composes task vectors in weight space, but the fused model is typically unstable and fails to match all teachers' full capability. In short, these methods trade off learning efficiency, attainable peak, and training stability against one another, leaving no satisfactory middle ground.

To overcome these limitations, we propose \textbf{Multi-teacher On-Policy Distillation} (\MOPD{}), a post-training paradigm that performs multi-teacher capability integration directly in policy space, rather than in weight space or in a static dataset space. \MOPD{} first conducts specialised RL independently on each domain to obtain a collection of strong domain teachers. Then the capabilities of these teachers are distilled into a single student model via on-policy distillation~\citep{hinton2015distilling, gu2024minillm, agarwal2023onpolicy}. This design has three structural benefits: (1) the training signal is dense and comes directly from each specialised teacher, so the student inherits the bulk of every teacher's capability; (2) training runs entirely on the student's own rollout distribution, which eliminates exposure bias by construction; (3) capability integration is realised in policy space through per-prompt routing rather than weight-space fusion, yielding a markedly more stable procedure.

We empirically validate \MOPD{} on two base models of different scales and architectures. On Qwen3-30B-A3B~\citep{qwen2025qwen3}, we compare \MixRL{}, \SeqRL{}, \RFT{}, \PM{}, and \MOPD{} across three domains (Math, Instruction Following, and Software Engineering~\citep{jimenez2024swebench, openai2024swebenchverified}); \MOPD{} leads the strongest baseline by $5.5$ points on the normalised score ($0.937$ vs.\ $0.882$). We further deploy \MOPD{} on the industrial-scale model MiMo-V2-Flash~\citep{xiaomi2026mimov2flash}, demonstrating its practical effectiveness in frontier-scale settings. Further analysis verifies the top-$k$ distillation instantiation, confirms that same-origin teachers are critical for stable optimization, and demonstrates the benefits of multi-round iterative refinement.

Our contributions are summarised as follows. (i) We propose \MOPD{}, a post-training paradigm that effects multi-teacher capability integration through dense, token-level supervision delivered on the student's own rollouts. (ii) We validate \MOPD{} on Qwen3-30B-A3B, showing that it outperforms existing capability-integration baselines, and further apply it to the industrial-scale MiMo-V2-Flash to confirm practical viability.

\afterpage{\afterpage{\begin{figure}[!t]
  \centering
  \includegraphics[width=\linewidth]{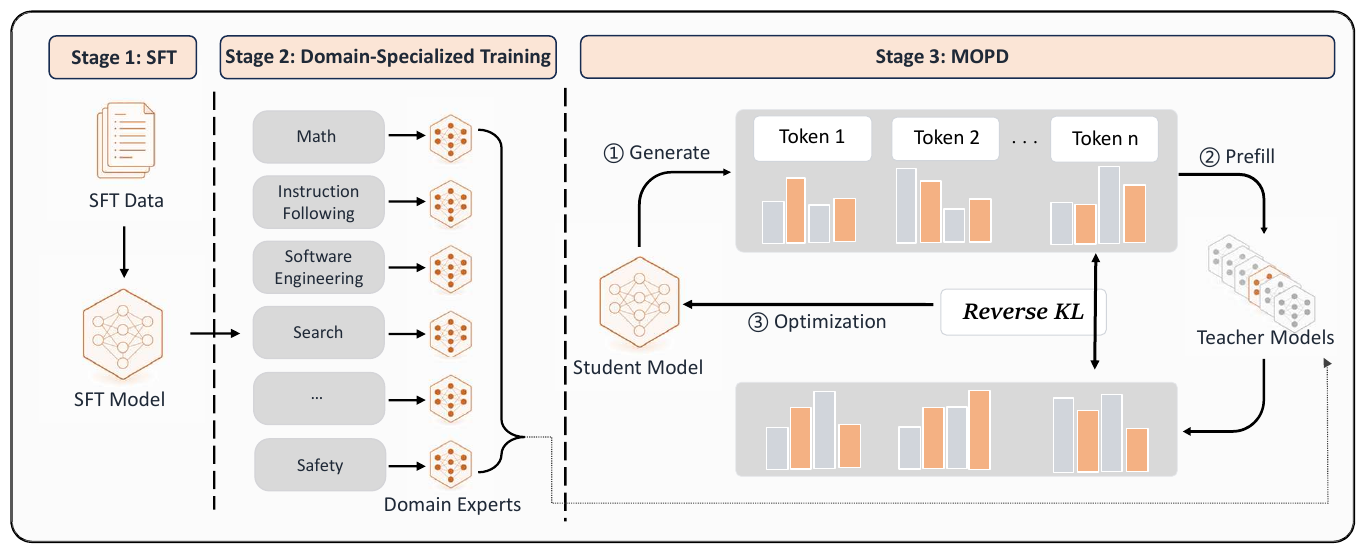}
  \caption{Overview of the three-stage \MOPD{} pipeline. Stage~1 produces a shared SFT checkpoint; Stage~2 trains the per-domain teachers in parallel; Stage~3 repeatedly (i)~samples a prompt from the domain mixture, (ii)~routes it to the matching teacher, (iii)~has the student roll out, (iv)~prefills the teacher on the rollout to obtain token-level log-probabilities, and (v)~updates the student using the teacher signal.}
  \label{fig:overview}
\end{figure}
}}

\section{Related Work}
\label{sec:related}

\paragraph{Reinforcement learning for LLM.}
In reinforcement learning with verifiable rewards (RLVR), PPO-style~\citep{schulman2017ppo} and GRPO-style~\citep{shao2024deepseekmath} outcome-reward optimization has become standard practice. These RL pipelines span a diverse task landscape, including math reasoning~\citep{guo2025deepseekr1, shao2024deepseekmath}, software engineering~\citep{wei2025swerl, jain2025r2e}, instruction following and creative writing~\citep{ouyang2022instructgpt}, and search agents~\citep{nakano2021webgpt, jin2025searchr1}. Each such pipeline yields a strong domain expert, and consolidating them into a single model naturally raises the multi-domain / multi-task question. For this setting, prior work mainly takes two routes: mixed RL~\citep{qwen2025qwen3}, which puts samples from different domains into the same training batch (each sample still uses its own per-domain reward and advantage), and cascade RL~\citep{wang2026nemotroncascadescalingcascadedreinforcement}, which trains the domains one after another.

\paragraph{Distillation for LLMs.}
Classical distillation~\citep{hinton2015distilling, kim2016seqkd} minimises a forward KL on a fixed corpus of teacher completions and is thus \emph{off-policy}, so the student's inference-time trajectories can drift from the training distribution. On-policy distillation addresses this by training on the student's own rollouts with the teacher as a scoring signal, as in MiniLLM~\citep{gu2024minillm} (reverse KL) and teacher-as-reward variants~\citep{agarwal2023onpolicy, meng2024simpo}. However, it is restricted to a single teacher and a single domain. \MOPD{} keeps the ``student samples, teacher scores'' template and extends it to multiple, prompt-routed domain teachers, validated on production-scale base models.

\paragraph{Model merging.}
Model merging fuses multiple model checkpoints in weight space to obtain a combined model without additional training. Model Soups~\citep{wortsman2022soups} averages the weights of independently fine-tuned models from the same initialisation. Task arithmetic~\citep{ilharco2023arithmetic} composes capabilities by adding and subtracting task vectors in weight space, while subsequent work mitigates parameter conflicts via TIES-Merging~\citep{yadav2023ties}, DARE~\citep{yu2024dare}, and AdaMerging~\citep{yang2024adamerging}.

\section{Method}
\label{sec:method}

\subsection{Pipeline}
\label{sec:method-pipeline}

\MOPD{} structures post-training as three sequential stages: general SFT, domain-specialised RL training, and a final \MOPD{} pass that fuses the per-domain experts into a single model (Fig.~\ref{fig:overview}).

\begin{itemize}[leftmargin=1.4em,itemsep=2pt,topsep=2pt,parsep=0pt]
  \item \textbf{Stage 1: General SFT.} We first fine-tune the base model on a broad corpus that covers all the base capabilities we care about, yielding an SFT checkpoint. This checkpoint serves as the shared initialisation for the domain-specialised RL training in Stage 2. And it is also used as the initialisation for the student of MOPD in Stage 3.
  \item \textbf{Stage 2: Domain-specialised RL training.} In this stage, for each domain $d$, we train a domain expert $\pi_{\phi_d}$ starting from the Stage-1 checkpoint. Each expert is trained with the RL recipe most natural for its domain, e.g., verifiable-answer RL for math and executable-sandbox agent RL for software engineering. The per-domain training are independent and can be executed fully in parallel.
  \item \textbf{Stage 3: \MOPD{}.} The student is initialised from the Stage-1 SFT checkpoint, and the Stage-2 domain experts $\{\pi_{\phi_d}\}$ are frozen and form the teacher group; then we train the student on a multi-domain dataset, where each optimization step performs the following operations: (1) we sample a batch of prompts from the dataset; (2) the student generates a trajectory for each prompt and records its per-token probability distribution along the trajectory; (3) each trajectory is dispatched to its corresponding domain teacher according to the task's domain, and then the teacher prefills on the trajectory to obtain its per-token probability distribution; (4) the student is updated by minimising the per-token reverse KL between the student and the domain teacher along the trajectory. After training, Stage 3 yields a single unified model that has acquired capabilities across all domains.
\end{itemize}

\MOPD{} offers several structural advantages over existing capability-integration paradigms:
\begin{itemize}[leftmargin=1.4em,itemsep=2pt,topsep=2pt,parsep=0pt]
  \item \textbf{No exposure bias.} The student is trained on its own rollouts, so the training and inference state distributions match by construction.
  \item \textbf{Dense per-token supervision.} The teacher provides a probability distribution at every token of the trajectory. This is far denser than the trajectory-level reward used in standard RL, which lowers variance and improves sample efficiency.
  \item \textbf{Stable integration in policy space.} Capabilities are merged by routing each prompt to its matching teacher, not by averaging or task arithmetic on teacher parameters. This avoids the instability typical of weight-space fusion.
  \item \textbf{Modular and parallel teachers.} Teacher RL training runs fully in parallel, and each teacher can use its own hyperparameter configuration. An unexpected failure or hyperparameter adjustment in one teacher's RL training does not affect the training of other teachers.
  \item \textbf{Same-origin teacher stability.} Because each teacher is produced by domain-specific RL from the same SFT checkpoint that initialises the student, teacher and student share a closely aligned policy distribution. This ensures low initial KL divergence and smooth, stable optimization throughout distillation (\S\ref{sec:exp-distribution}).
\end{itemize}

\subsection{Algorithm}
\label{sec:method-algorithm}
\label{sec:method-core}

In Stage 3, the student is optimised against a per-token reverse KL toward the dispatched teacher:
\begin{equation}
  \mathcal L_{\mathrm{rev\text{-}KL}} = \mathbb E_{x,\, y \sim \pi_\theta}\!\Big[ \tfrac{1}{|y|}\sum_t \sum_{v} \pi_\theta(v) \log \tfrac{\pi_\theta(v)}{\pi_{\phi_d}(v)} \Big],
  \label{eq:revkl}
\end{equation}
where $\pi_{\phi_d}$ is the teacher dispatched to prompt $x$, and $\pi_\theta(v)$, $\pi_{\phi_d}(v)$ are shorthand for $\pi_\theta(v \mid x, y_{<t})$, $\pi_{\phi_d}(v \mid x, y_{<t})$ (same shorthand applied below). We give two efficient implementations of this objective in the next two subsections.

\subsubsection{Policy-gradient implementation}
\label{sec:method-pg}

Following MiniLLM~\citep{gu2024minillm}, we cast \MOPD{} distillation as an RL process. The gradient of \eqref{eq:revkl} is
\begin{equation}
  \nabla_\theta \mathcal L = -\mathbb E_{x,y}\!\Big[ \tfrac{1}{|y|}\sum_t \log \tfrac{\pi_{\phi_d}(y_t)}{\pi_\theta(y_t)} \nabla_\theta \log \pi_\theta(y_t) \Big],
  \label{eq:revkl-grad}
\end{equation}
which is exactly the policy-gradient form, with the teacher--student log-difference acting as a per-token \emph{advantage}:
\begin{equation}
  \hat A_{\MOPD, t} = \sg[\, \log \pi_{\phi_d}(y_t) - \log \pi_\theta(y_t) \,].
  \label{eq:advantage}
\end{equation}
For training stability, we two-side-clip the advantage, $\hat A^{\mathrm{clip}}_{\MOPD, t} = \mathrm{clip}(\hat A_{\MOPD, t}, -A_{\max}, +A_{\max})$, giving the final loss
\begin{equation}
  \mathcal L^{\mathrm{PG}}_{\MOPD}(\theta) = -\mathbb E_{x,y}\!\Big[ \tfrac{1}{|y|}\sum_t \hat A^{\mathrm{clip}}_{\MOPD, t}\, \log \pi_\theta(y_t) \Big].
  \label{eq:mopd-loss}
\end{equation}
This form drops directly into existing PPO/GRPO training frameworks; the advantage computation is the only change.

\subsubsection{Top-$k$ distillation implementation}
\label{sec:method-topk}

The policy-gradient form uses only the single sampled token at each rollout position. A lower-variance alternative is to distill on the teacher's top-$k$ tokens~\citep{peng2024pretrainingdistillation}, which exploits more of the teacher's distribution while keeping the implementation simple. Let $\mathcal T^{d}_t = \mathrm{TopK}_k(\pi_{\phi_d}(\cdot \mid x, y_{<t}))$ be the top-$k$ token set under the teacher at position $t$. We use the following distillation loss:
\begin{equation}
  \mathcal L^{\mathrm{TopK}}_{\MOPD}(\theta) = \mathbb E_{x,y}\!\Big[ \tfrac{1}{|y|}\sum_t \sum_{v \in \mathcal T^{d}_t} \!\big[\, \pi_\theta(v) \log \tfrac{\pi_\theta(v)}{\pi_{\phi_d}(v)} - \pi_\theta(v) + \pi_{\phi_d}(v) \big] \Big],
  \label{eq:mopd-topk}
\end{equation}
where $\pi_\theta(v)$ and $\pi_{\phi_d}(v)$ are both conditioned on $(x, y_{<t})$; we drop the conditioning for brevity.

We have added an extra $\pi_{\phi_d}(v) - \pi_\theta(v)$ term beyond the standard reverse KL. This added term corrects the bias introduced by top-$k$ truncation, ensuring that the loss is minimised at $\pi_\theta = \pi_{\phi_d}$ on the top-$k$ tokens; without it, the naive top-$k$-truncated reverse KL does not have this property.

Beyond correctness, the top-$k$ form also reduces the infrastructure burden: it keeps the teacher-prefill payload small enough to be transmitted like a lightweight reward signal, in contrast to full-vocabulary distillation which requires shipping the entire vocabulary distribution per token.

\subsection{Infrastructure}
\label{sec:method-infra}

The additional operation in Stage 3 is the teacher prefill: every student rollout has to be passed through the dispatched teacher to obtain per-token log-probabilities (or top-$k$ logits). A natural baseline is to fold teacher prefill into the RL training loop, but this complicates the infrastructure and adds extra serial latency. We observe that the teacher prefill computation has the same nature as reward computation in RL, so we instead deploy each domain teacher as a stand-alone prefill service sitting outside the RL trainer.

The runtime works as follows. The student sampler keeps generating rollouts; as soon as one sequence's rollout finishes, the trainer fires an asynchronous prefill request to the corresponding teacher service. The teacher service computes and returns the per-token log-probabilities. Because teacher prefill and other sequences' sampling overlap in time, the wall-clock cost of \MOPD{} is dominated by sampling alone, and the teacher cost is essentially hidden behind student sampling. In our deployments, the teacher contributed nearly no measurable wall-clock overhead.

\section{Experiments}
\label{sec:experiments}

We compare \MOPD{} against four capability-integration paradigms on Qwen3-30B-A3B under controlled conditions (\S\ref{sec:exp-qwen3-setup}--\ref{sec:exp-qwen3-main}) and validate it on the industrial-scale frontier model MiMo-V2-Flash (\S\ref{sec:exp-mimo}). Finally, \S\ref{sec:exp-extensions} presents our analysis.

\subsection{Experimental setup}
\label{sec:exp-qwen3-setup}

\paragraph{SFT model.}
Our starting checkpoint is Qwen3-30B-A3B-Base~\citep{qwen2025qwen3}. To make it a strong starting point for RL, we fine-tune it on a broad-coverage corpus spanning all target domains. All baselines and \MOPD{} start from this same SFT checkpoint.

\paragraph{Tasks and evaluation.}
Our tasks span three domains: math (evaluated on AIME25 and AIME26), instruction following (evaluated on IFBench~\citep{pyatkin2025ifbench} and IFEval~\citep{zhou2023ifeval}), and software engineering (evaluated on SWE-bench Verified~\citep{jimenez2024swebench}). Training data sources and hyperparameters are described in Appendix~\ref{app:training-details}.

\paragraph{Baselines.}
We compare five capability-integration paradigms: (i) \MixRL{}, joint RL on a dataset that pools prompts from every domain; (ii) \SeqRL{}, per-domain RL applied sequentially with each stage initialised from the previous; (iii) \RFT{}, offline SFT of the student on rollouts produced by the Stage-2 teachers; (iv) \PM{}, linear / task-arithmetic merge of the Stage-2 teachers; and (v) \MOPD{} (ours).

\paragraph{Evaluation metric.}
The three domains differ substantially in absolute headroom, so a raw accuracy average over-weights the wider-headroom domains. To put the domains on a common scale we report a normalised score. Let $\mathcal D$ denote the set of evaluation domains. For each $d \in \mathcal D$, let $s_d^{\text{s}}$ be the Stage-1 SFT student accuracy on that domain and $s_d^{\text{t}}$ the corresponding Stage-2 per-domain specialist accuracy; for a method with domain accuracy $s_d$, the per-domain normalised score is $\tilde s_d = (s_d - s_d^{\text{s}}) / (s_d^{\text{t}} - s_d^{\text{s}})$, which equals $0$ at the Stage-1 student and $1$ at the per-domain specialist teacher. The headline figure we report is the uniform average $\bar{\tilde s} = \tfrac{1}{|\mathcal D|}\sum_{d \in \mathcal D} \tilde s_d$ across domains; values above $1$ indicate improvement beyond the specialist teacher, values below $0$ indicate a regression from the Stage-1 student.

\subsection{Main results}
\label{sec:exp-qwen3-main}

\begin{table}[t]
  \centering
  \caption{Qwen3-30B-A3B: multi-domain capability integration. Per-benchmark columns are accuracy (\%); higher is better. ``RL Teacher'' is the Stage-2 specialised RL teacher for that domain. The \textbf{Norm.\ score} column is the normalised score defined in \S\ref{sec:exp-qwen3-setup}; higher is better. Within the six integration methods we \textbf{bold} the column best and \underline{underline} the runner-up.}
  \label{tab:qwen3-main}
  \small
  \setlength{\tabcolsep}{3pt}
  \begin{tabular}{lccccc|c}
    \toprule
    & \multicolumn{2}{c}{Math} & \multicolumn{2}{c}{Instruction Following} & SWE & \\
    \cmidrule(lr){2-3} \cmidrule(lr){4-5} \cmidrule(lr){6-6}
    Method & AIME25 & AIME26 & IFBench & IFEval & SWE-bench Verified & \makecell{Norm.\\score} \\
    \midrule
    Student (SFT-only)   & 45.42 & 54.48 & 42.69 & 84.17 & 35.80 & 0.0000 \\
    RL Teacher               & 54.79 & 63.65 & 78.40 & 95.50 & 51.20 & 1.0000 \\
    \midrule
    \MixRL{}             & \textbf{52.71} & 63.75 & 75.00 & 94.58 & \underline{48.80} & \underline{0.8818} \\
    \SeqRL{}             & 48.54 & 61.88 & 77.11 & \underline{95.80} & 47.80 & 0.7752 \\
    \RFT{}               & \underline{51.56} & 63.44 & \textbf{80.95} & 93.35 & 45.80 & 0.8241 \\
    \PM{} (Avg.)         & 47.81 & 59.58 & 53.74 & 88.79 & 39.60 & 0.3280 \\
    \PM{} (Task Arith.)  & 49.38 & \underline{63.96} & \underline{78.23} & \textbf{95.81} & \underline{48.80} & 0.8574 \\
    \midrule
    \MOPD{} (ours)       & 51.46 & \textbf{65.31} & 77.89 & 93.84 & \textbf{50.40} & \textbf{0.9373} \\
    $\Delta$(\MOPD{} $-$ RL Teacher) & $-3.33$ & $+1.66$ & $-0.51$ & $-1.66$ & $-0.80$ & $-0.0627$ \\
    \bottomrule
  \end{tabular}
\end{table}

\begin{figure}[t]
  \centering
  \includegraphics[width=\linewidth]{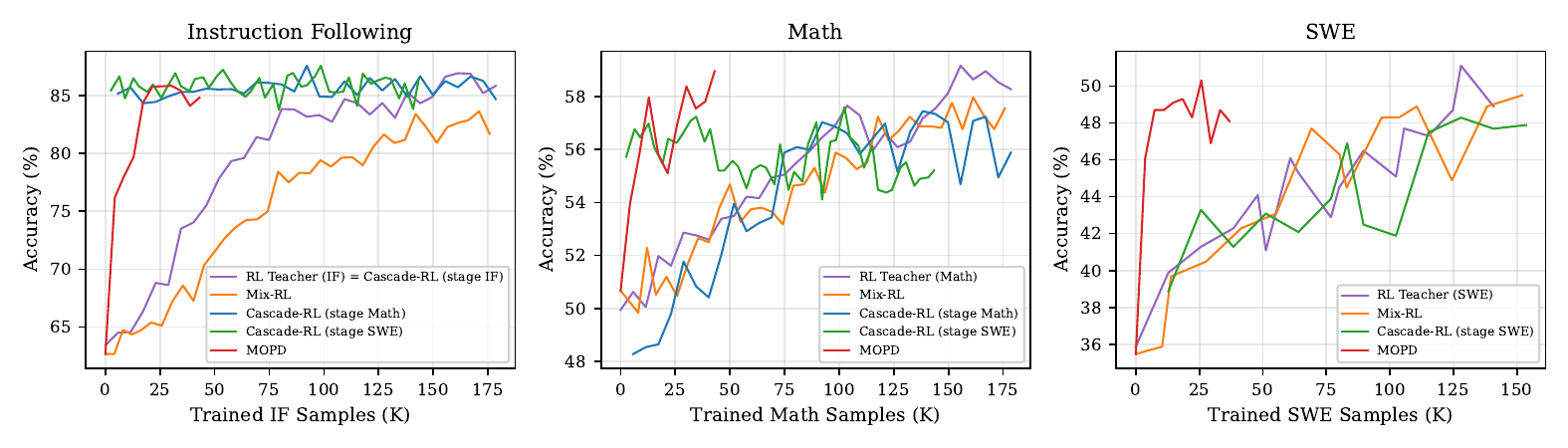}
  \caption{Training dynamics on Qwen3-30B-A3B. Each panel is a domain-level accuracy (\%); the $x$-axis is cumulative training samples (K). Domain-RL (purple) is the single-domain specialist teacher for reference.}
  \label{fig:training-curves}
\end{figure}

Table~\ref{tab:qwen3-main} reports per-benchmark accuracy and the normalised score across all three domains. Figure~\ref{fig:training-curves} tracks domain-level accuracy against per-domain training samples. We make four observations.

\begin{itemize}[leftmargin=*]
\item \textbf{\MixRL{}, \SeqRL{}, and Off-Policy Finetune each leave domain-specific gaps, in different shapes.}
The three methods reach normalised scores of $0.882$, $0.775$, and $0.824$ respectively, similar in aggregate but with distinct per-domain weaknesses. \SeqRL{} (trained in order IF$\to$Math$\to$SWE) exhibits cross-domain interference: the first-trained IF domain closes $98\%$ of the headroom, but Math (second stage) closes only $57\%$; Figure~\ref{fig:training-curves} further shows that Math accuracy degrades during the subsequent SWE stage. \RFT{} exceeds the teacher on IF (per-domain normalised score $1.01$) yet closes only $65\%$ on SWE, suggesting that offline imitation of teacher trajectories transfers unevenly across task types. \MixRL{} is the most balanced baseline (per-domain range $0.064$) but still trails \MOPD{} by $5.5$ points overall.

\item \textbf{\MOPD{} achieves the highest normalised score with the most uniform profile.}
\MOPD{} reaches a normalised score of $0.937$, $+0.055$ over the next-best method (\MixRL{} at $0.882$). Its three per-domain normalised scores fall in $[0.91, 0.95]$, a range of only $0.044$, the smallest of any method. No other approach closes the headroom this uniformly: \SeqRL{} spans $0.57$--$0.98$ (range $0.41$) and \RFT{} spans $0.65$--$1.01$ (range $0.36$).

\item \textbf{\PM{} is highly sensitive to the merging recipe.}
Linear averaging fails broadly (normalised score $0.328$); Task Arithmetic recovers to $0.857$, but its per-domain profile varies widely: it exceeds the teacher on IF ($1.00$) yet closes only $73\%$ on Math. The outcome depends on both the merging coefficients and the benchmark, making parameter merging an unreliable integration tool.

\item \textbf{\MOPD{} is markedly more sample-efficient.}
As Figure~\ref{fig:training-curves} shows (x-axis: per-domain samples consumed), \MOPD{} reaches the teacher-level plateau on IF within ${\sim}25$K IF samples and on SWE within ${\sim}30$K SWE samples, whereas \MixRL{} requires the full $150$--$180$K sample budget \emph{in each domain} to approach a comparable level. The dense per-token supervision from the teacher provides richer gradient signal per sample than trajectory-level RL reward, explaining the faster convergence.
\end{itemize}

\subsection{Scaling to a 309B-parameter model}
\label{sec:exp-mimo}
\begin{table}[t]
  \centering
  \caption{MiMo-V2-Flash: integration via the full \MOPD{} pipeline. All teachers are RL-trained.}
  \label{tab:mimo-main}
  \small
  \setlength{\tabcolsep}{3.5pt}
  \begin{tabular}{lccccccc}
    \toprule
     & AIME25 & HMMT25 & LCB & IFBench & SWE-Bench V. & $\tau^2$-Bench & $\tau^2$-Telecom \\
    \midrule
    Student   & 89.3 & 76.9 & 77.5 & 55.4 & 67.8 & 75.9 & 92.7 \\
    Teacher   & 93.9 & 82.6 & 82.6 & 68.9 & 74.2 & 79.6 & 95.0 \\
    \MOPD{}   & 94.1 & 84.4 & 83.2 & 66.7 & 73.4 & 80.3 & 95.3 \\
    $\Delta$  & $+0.2$ & $+1.8$ & $+0.6$ & $-2.2$ & $-0.8$ & $+0.7$ & $+0.3$ \\
    \bottomrule
  \end{tabular}
\end{table}

To confirm that \MOPD{} carries over to industrial-scale frontier models, we apply the full three-stage pipeline to MiMo-V2-Flash. We use domain teachers covering Math, Code, IF, SWE, and Tool Use.

Table~\ref{tab:mimo-main} presents the \MOPD{} results on MiMo-V2-Flash. \MOPD{} matches or exceeds the corresponding teacher on most benchmarks. The two regressions, IFBench ($-2.2$) and SWE-Bench Verified ($-0.8$), are modest relative to the gains on the remaining benchmarks.

\subsection{Analysis}
\label{sec:exp-extensions}

\subsubsection{Policy-gradient and top-$k$ distillation perform comparably}
\label{sec:exp-distill-variant}
\afterpage{%
  \begin{table}[t]
  \centering
  \caption{Distillation loss variants and teacher distribution alignment on Qwen3-30B-A3B. Per-benchmark columns are accuracy (\%); the last column is the normalised score defined in \S\ref{sec:exp-qwen3-setup}. The top two rows use the RL teachers from the default \MOPD{} pipeline; the bottom two rows replace the Math teacher with Qwen3-235B-A22B, a stronger but distributionally different model. Both loss variants use $k=64$ for top-$k$.}
  \label{tab:distill-variant}
  \small
  \setlength{\tabcolsep}{5pt}
  \resizebox{\linewidth}{!}{%
  \begin{tabular}{ll cccccc}
    \toprule
    & & \multicolumn{2}{c}{Math} & \multicolumn{2}{c}{Instruction Following} & SWE & \\
    \cmidrule(lr){3-4} \cmidrule(lr){5-6} \cmidrule(lr){7-7}
    Math Teacher & Loss Variant & AIME25 & AIME26 & IFBench & IFEval & \makecell{SWE-bench\\Verified} & \makecell{Norm.\\score} \\
    \midrule
    \multirow{2}{*}{RL Teacher}
      & Policy gradient & 51.46 & 65.31 & 77.89 & 93.84 & 50.40 & 0.9373 \\
      & Top-$k$ distillation & 51.77 & 64.79 & 75.85 & 93.07 & 50.20 & 0.9093 \\
    \midrule
    \multirow{2}{*}{Qwen3-235B-A22B}
      & Policy gradient & 45.63 & 51.56 & 79.25 & 93.99 & 50.60 & 0.6003 \\
      & Top-$k$ distillation & 0.94 & 0.42 & 72.96 & 88.97 & 51.20 & $-1.1898$ \\
    \bottomrule
  \end{tabular}%
  }
\end{table}
  \begin{figure}[t]
  \centering
  \includegraphics[width=\linewidth]{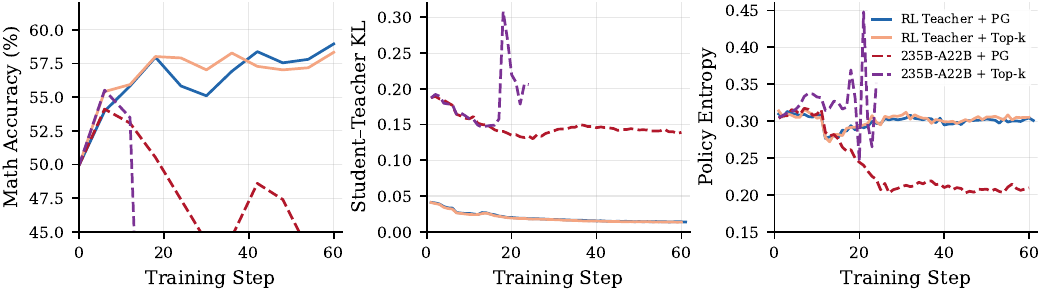}
  \caption{Training dynamics across the four runs in Table~\ref{tab:distill-variant} on the Math domain of Qwen3-30B-A3B. \textbf{Left}: Math accuracy (\%). \textbf{Middle}: per-token student--teacher reverse KL divergence. \textbf{Right}: student policy entropy. The two same-origin teacher runs (solid) show smooth, stable optimization; the two external-teacher runs (dashed) exhibit progressively unstable training, with the top-$k$ variant collapsing catastrophically around step~18.}
  \label{fig:distill-variant-curves}
\end{figure}
}

\S\ref{sec:method-algorithm} introduced two instantiations of the \MOPD{} distillation loss: the policy-gradient form \eqref{eq:mopd-loss} and the top-$k$ form \eqref{eq:mopd-topk}. We compare the two on Qwen3-30B-A3B under an otherwise identical pipeline (same teachers, data budget, and hyperparameters), with $k=64$ for the top-$k$ variant.

The top two rows of Table~\ref{tab:distill-variant} report the comparison. Top-$k$ distillation is comparable on Math and slightly worse on IF and SWE; the two forms reach overall similar capability, with normalised scores of $0.909$ vs.\ $0.937$. Figure~\ref{fig:distill-variant-curves} corroborates this: under same-origin teachers, both loss forms produce nearly overlapping training trajectories: math accuracy climbs smoothly, the per-token reverse KL decreases monotonically from an already low initial value (${\sim}0.04$), and policy entropy remains stable around $0.30$ throughout. When teacher and student share a closely related distribution, the student's rollouts concentrate in the teacher's high-probability region, so both gradient estimators receive similar informative signal and converge to comparable endpoints. This indicates that, with same-origin teachers, the policy-gradient form is already sufficiently stable; introducing top-$k$ does not yield additional stability or variance reduction.

\subsubsection{Same-origin teachers stabilize distillation}
\label{sec:exp-distribution}

\MOPD{}'s domain teachers are obtained by RL training from the student. This means that teachers and the student share a similar policy distribution. Recent studies~\citep{li2026rethinkingonpolicydistillation, ko2026scalingreasoningrelaxed} discuss the relationship between teacher--student policy divergence and on-policy distillation stability. A natural question arises: would replacing a teacher with a stronger but distributionally different external model yield better results? To test this, we design a controlled experiment: in the Math domain of \MOPD{}, we replace the original Qwen3-30B-A3B Math RL teacher with Qwen3-235B-A22B, a significantly larger model with stronger math capability. All other settings (IF and SWE teachers, data, hyperparameters) remain unchanged.

The bottom two rows of Table~\ref{tab:distill-variant} report the results. Despite Qwen3-235B-A22B's substantially stronger absolute math capability, the \MOPD{} student's Math performance degrades under both loss variants. Figure~\ref{fig:distill-variant-curves} reveals the mechanism. The initial per-token KL is approximately $5{\times}$ higher than the same-origin setting (${\sim}0.19$ vs.\ ${\sim}0.04$), quantitatively confirming the large distributional gap. Under the policy-gradient form, math accuracy degrades progressively while entropy contracts from $0.30$ to $0.21$; the student's policy narrows toward a single mode as it receives predominantly punitive gradient signal from the teacher's low-probability region. Under the top-$k$ form, the collapse is even more severe: training diverges catastrophically around step~18, with wild oscillation in both KL and entropy, indicating a complete loss of optimization stability. These results underscore that \MOPD{}'s use of same-origin teachers is critical: the close distributional alignment ensures stable optimization throughout training, yielding reliable and consistent gains.

\subsubsection{Multi-round student--teacher evolution}
\label{sec:exp-coevolve}
\label{sec:method-extensions}
\label{sec:method-ext}

A single round of \MOPD{} closes most of the student--teacher headroom on every domain, but Table~\ref{tab:coevolve} shows that headroom remains in the Math and IF domains. We propose using the post-\MOPD{} model as a new student and repeating the procedure: retrain each per-domain teacher from this student, then perform another \MOPD{} integration with the new teachers.

We validate one such follow-up round on Qwen3-30B-A3B. After the first round, the student becomes the initialisation for the next round's per-domain RL teachers, which drive the second \MOPD{} pass. In this round, we apply RL training and \MOPD{} distillation to the Math and IF domains only (for SWE, we perform neither RL training nor distillation).

\begin{table}[t]
  \centering
  \caption{Multi-round student--teacher evolution on Qwen3-30B-A3B. Per-benchmark columns are accuracy (\%); the last column is the normalised score defined in \S\ref{sec:exp-qwen3-setup}. Iter-2 RL Teacher is the per-domain RL teacher initialised from the Iter-1 \MOPD{} student; Iter-2 \MOPD{} is the student produced from those teachers. No new SWE teacher was trained at Iter~2; the SWE scores reflect only indirect effects of Math/IF distillation.}
  \label{tab:coevolve}
  \small
  \setlength{\tabcolsep}{3pt}
  \begin{tabular}{lcccccc}
    \toprule
    & \multicolumn{2}{c}{Math} & \multicolumn{2}{c}{Instruction Following} & SWE & \\
    \cmidrule(lr){2-3} \cmidrule(lr){4-5} \cmidrule(lr){6-6}
    Round & AIME25 & AIME26 & IFBench & IFEval & SWE-bench Verified & \makecell{Norm.\\score} \\
    \midrule
    Iter~1 \MOPD{}      & 51.46 & 65.31 & 77.89 & 93.84 & 50.40 & 0.937 \\
    Iter~2 RL Teacher   & 54.27 & 65.52 & 81.46 & 95.65 & 50.40 & 1.030 \\
    Iter~2 \MOPD{}      & 53.44 & 64.90 & 79.76 & 95.44 & 50.20 & 0.986 \\
    \bottomrule
  \end{tabular}
\end{table}

We observe: (i) initialising the next round's per-domain RL from the Iter-1 student does produce stronger teachers: the Iter-2 RL Teacher reaches a normalised score of $1.030$, $+0.093$ over Iter-1 \MOPD{}; this confirms that per-domain capability headroom remains beyond what a single \MOPD{} round extracts. (ii) Guided by the Iter-2 RL teachers, the Iter-2 student improves further on Math and IF, with its normalised score rising from $0.937$ to $0.986$ ($+0.049$), showing that the \MOPD{} pipeline can continuously absorb capability from stronger teachers.

\section{Discussion}
\label{sec:discussion}

Beyond its empirical gains, \MOPD{} structurally decouples the serial dependencies intrinsic to multi-domain post-training, separating capability production (Stage-2 per-domain RL) from capability integration (Stage-3 distillation). Three workflow consequences follow:
\begin{itemize}[leftmargin=1.4em,itemsep=2pt,topsep=2pt,parsep=0pt]
  \item \textbf{Parallel development.} Stage-2 teachers are mutually independent: each domain team iterates on its own rewards, sandboxes, and data pipelines concurrently, with no fixed ordering. Given sufficient compute, this directly raises the team's overall development throughput.
  \item \textbf{Recipe-level decoupling.} Each domain is free to choose its own RL recipe (algorithm, rollout procedure, reward function, hyperparameters, and so on) without worrying about conflicts with other domains or about its own optimization choices interfering with theirs.
  \item \textbf{Risk isolation.} RL training often requires iterative tuning of algorithms and hyperparameters (whether to push capability higher or simply to address stability issues), and each such adjustment forces a training restart. Under joint multi-domain RL, a restart sends the entire run back to the start; under \MOPD{}'s parallel teacher development, a restart is confined to the single affected domain and the other teachers continue undisturbed.
\end{itemize}

\section{Conclusion}
\label{sec:conclusion}

We proposed \MOPD{} (Multi-Teacher On-Policy Distillation), a post-training paradigm that performs multi-domain capability integration in policy space: the student samples from its own rollouts, each prompt is routed to the corresponding frozen domain teacher, and the teacher provides a dense, token-level log-probability signal at every position. On Qwen3-30B-A3B, \MOPD{} attains the best aggregate integration among \MixRL{}, \SeqRL{}, \RFT{}, and \PM{}, leading the next-best integration method by $5.5$ normalised-score points while closing $91$--$95\%$ of the student--teacher headroom on every domain; on MiMo-V2-Flash, the same recipe matches or exceeds the corresponding teacher on most benchmarks. Our analysis further shows that both policy-gradient and top-$k$ loss forms perform comparably, that same-origin teachers are critical for stable optimization, and that the pipeline supports multi-round student--teacher evolution. Combined with the structural decoupling discussed in \S\ref{sec:discussion}, we believe \MOPD{} offers an engineering path for LLM post-training at scale that supports parallel development and rapid integration.

\bibliography{references,custom}

\appendix
\clearpage
\section{Training Details}
\label{app:training-details}

\paragraph{Training data.}
Our training tasks span three domains:
\begin{enumerate}[leftmargin=1.6em,itemsep=1pt,topsep=2pt,parsep=0pt]
  \item \textbf{Math}. SFT data come primarily from the math subset of the Mixture-of-Thoughts dataset~\citep{openr1_mot2025}; RL training data are collected and filtered from many open-source datasets, including BigMath~\citep{albalak2025bigmath} and ORZ~\citep{hu2025orz}. The maximum sequence length is 32{,}768.
  \item \textbf{Instruction following}. SFT prompts are constructed following the method introduced by IFBench~\citep{pyatkin2025ifbench} and distilled on gpt-oss-120b~\citep{openai2025gptoss120b}; RL training data are similarly synthesised following the IFBench recipe. The maximum sequence length is 32{,}768.
  \item \textbf{Software engineering}. SFT data are distilled from R2E-Gym tasks via open-source models; RL training data use R2E-Gym-Lite~\citep{jain2025r2e}. The maximum sequence length is 65{,}536 and the number of interaction turns is capped at 50.
\end{enumerate}

\paragraph{Hyperparameters.}
For RL training, we use an on-policy GRPO~\citep{shao2024deepseekmath} algorithm with Dynamic Sampling~\citep{dapo}: data generated from each rollout is used for a single gradient update and then discarded. For per-domain RL, we set the learning rate to $3\times 10^{-6}$. For Math and IF training, we set batch size (BS) to $144$ with $N=8$ rollouts per prompt, training approximately $175$K sequences. For SWE, we set BS to $80$ with $N=8$, training approximately $150$K sequences. \SeqRL{} uses the same configuration as per-domain RL, with domains trained in the order IF$\to$Math$\to$SWE. For \MixRL{}, we set the learning rate to $4\times 10^{-6}$, BS $256$, $N=8$, with a per-batch domain ratio of Math\,:\,IF\,:\,SWE $= 0.35:0.35:0.3$. For \MOPD{}, we do not use Dynamic Sampling and set BS $2048$, $N=1$, with a per-batch domain ratio of Math\,:\,IF\,:\,SWE $= 0.35:0.35:0.3$. The default advantage clip maximum is $5$ for the policy-gradient form; the default $k$ is $64$ for the top-$k$ distillation form.

\section{Evaluation Details}
\label{app:eval-details}

We describe the evaluation protocol used for all reported results. For the two math benchmarks, AIME25 and AIME26, we sample $32$ times per question and report the average accuracy (avg@32) to reduce the variance inherent to low-sample-count, high-difficulty math evaluation. For instruction following, we evaluate each question once on both IFBench and IFEval. For software engineering, we evaluate each task once on SWE-bench Verified.

For all benchmarks, decoding uses a sampling temperature of $1.0$, with neither top-$p$ nor top-$k$ truncation applied.

\end{document}